\documentclass[10pt,twocolumn,letterpaper]{article}

\usepackage{cvpr}      

\usepackage[dvipsnames]{xcolor}

\definecolor{cvprblue}{rgb}{0.21,0.49,0.74}
\usepackage[pagebackref,breaklinks,colorlinks,citecolor=cvprblue]{hyperref}

\def\paperID{10797} 
\def\confName{CVPR}
\def\confYear{2024}

\usepackage[utf8]{inputenc} 
\usepackage[T1]{fontenc}    
\usepackage{hyperref}       
\usepackage{url}            
\usepackage{booktabs}       
\usepackage{amsfonts}       
\usepackage{nicefrac}       
\usepackage{microtype}      
\usepackage[accsupp]{axessibility}

\usepackage{graphicx}
\usepackage{amsmath}
\usepackage{booktabs}
\usepackage{stfloats}
\usepackage{multirow}
\usepackage{pifont}
\newcommand{\cmark}{\ding{51}}
\newcommand{\xmark}{\ding{55}}
\definecolor{DarkGreen}{rgb}{0.0, 0.5, 0.0}
\usepackage{makecell}

\usepackage{booktabs} 
\usepackage{multirow} 
\usepackage{xcolor}   
\usepackage{array}
\usepackage{colortbl} 

\definecolor{Gray}{gray}{0.9}

\title{ULIP-2: Towards Scalable Multimodal Pre-training for 3D Understanding}

\author{%
\textbf{Le Xue$^{1}$} \thanks{\ Contact: lxue@salesforce.com}
\hspace*{0.5em}
\textbf{Ning Yu$^{1}$} \hspace*{0.5em}
\textbf{Shu Zhang$^{1}$} \hspace*{0.5em}
\textbf{Artemis Panagopoulou$^{1,3}$} \hspace*{0.5em}
\textbf{Junnan Li$^{1}$} \hspace*{0.5em} \\
\textbf{Roberto Martín-Martín$^{4}$} \hspace*{0.5em}
\textbf{Jiajun Wu$^{2}$} \hspace*{0.5em} \\
\textbf{Caiming Xiong$^{1}$} \hspace*{0.5em}
\textbf{Ran Xu$^{1}$} \hspace*{0.5em}
\textbf{Juan Carlos Niebles$^{1,2}$} \hspace*{0.5em}
\textbf{Silvio Savarese$^{1, 2}$}
\\
$^{1}$ Salesforce AI Research \hspace*{0.1em} $^{2}$ Stanford University \hspace*{0.1em} \\
$^{3}$ University of Pennsylvania  \hspace*{0.1em} ${^4}$ University of Texas at Austin
}

\begin{document}

\newcommand{\customsize}{\fontsize{5}{8}\selectfont}

\newcommand{\customsizePointnext}{\fontsize{8}{11}\selectfont}

\newcommand{\customsizeTripletstats}{\fontsize{6}{9}\selectfont}

\maketitle

\begin{abstract}
\vspace{-2mm}
Recent advancements in multimodal pre-training have shown promising efficacy in 3D representation learning by aligning multimodal features across 3D shapes, their 2D counterparts, and language descriptions.
However, the methods used by existing frameworks to curate such multimodal data, in particular language descriptions for 3D shapes, are not scalable, and the collected language descriptions are not diverse. 
To address this, we introduce ULIP-2, a simple yet effective tri-modal pre-training framework that leverages large multimodal models to automatically generate holistic language descriptions for 3D shapes. It only needs 3D data as input, eliminating the need for any manual 3D annotations, and is therefore scalable to large datasets.
ULIP-2 is also equipped with scaled-up backbones for better multimodal representation learning.
We conduct experiments on two large-scale 3D datasets, Objaverse and ShapeNet, and augment them with tri-modal datasets of 3D point clouds, images, and language for training ULIP-2.
Experiments show that ULIP-2 demonstrates substantial benefits in three downstream tasks: zero-shot 3D classification, standard 3D classification with fine-tuning, and 3D captioning (3D-to-language generation).
It achieves a new SOTA of \textbf{50.6\% (top-1)} on Objaverse-LVIS and \textbf{84.7\% (top-1)} on ModelNet40 in zero-shot classification. In the ScanObjectNN benchmark for standard fine-tuning, ULIP-2 reaches an overall accuracy of \textbf{91.5\%} with a compact model of only 1.4 million parameters.
ULIP-2 sheds light on a new paradigm for scalable multimodal 3D representation learning without human annotations and shows significant improvements over existing baselines.
The code and datasets are released at https://github.com/salesforce/ULIP.

\end{abstract}

\section{Introduction}
\label{sec:intro}

\begin{figure}[h!]
    \centering
    \includegraphics[width=0.8\linewidth]{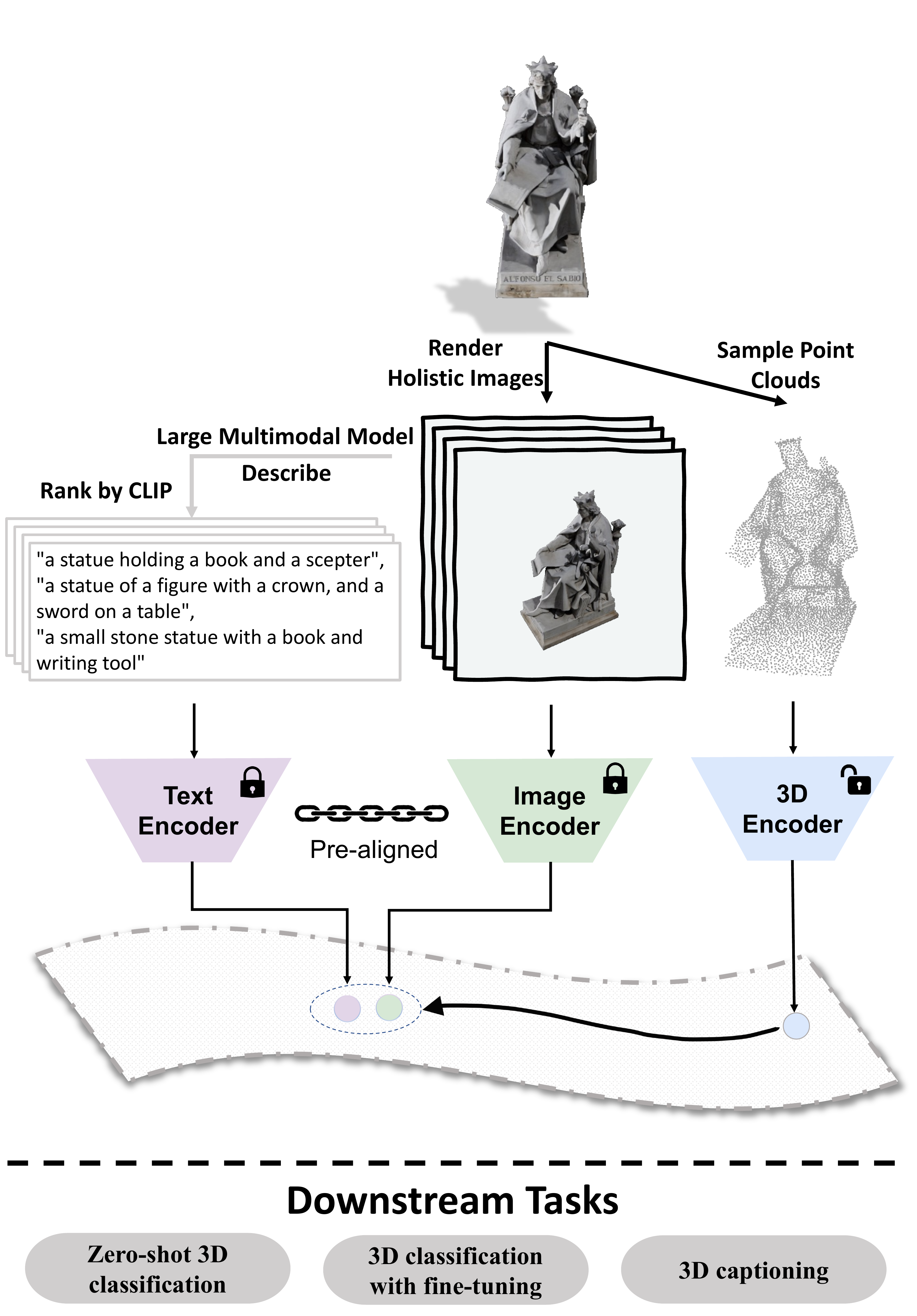}
    \caption{Overview of the ULIP-2 pre-training framework and its downstream tasks. The above part is the ULIP-2 pre-training framework, ULIP-2 employs a large multimodal model to automatically generate detailed descriptions for each 2D-rendered image from holistic viewpoints of a 3D shape. ULIP-2 takes advantage of a pre-aligned and frozen vision-language feature space to achieve alignment among the triplet modalities: holistic texts, images, and 3D point clouds. After the pre-training, the 3D encoder will be used in the downstream tasks. As shown in the figure, only the 3D data is required for this pre-training process.}
    
    \label{fig:overview}
     \vspace{-5mm}
\end{figure}

\vspace{-3mm}
3D visual understanding has seen a surge of interests in recent years \cite{hu2020randla,graham20183d,li2021lidar,liu2019densepoint, xu2023pointllm, qi2023gpt4point} due to its growing applications in augmented reality and virtual reality (AR and VR) \cite{armeni20163d,liu2021group,misra2021end,vu2022softgroup, shu2023model}, autonomous driving \cite{li2022deepfusion,yin2021center}, and robotics \cite{cadena2016multi,wojek2011monocular}. Despite this, the collections and annotations of 3D data remain a costly and labor-intensive process \cite{chang2015shapenet, uy2019revisiting, wu20153d}. In response to this challenge, researchers have turned to other more abundantly available modalities, \emph{e.g.,} image and natural language, to provide supervisory signals for learning 3D representations. This approach has not only led to improved unimodal representation but also cultivated a richer multimodal representation capability. The results have been promising, and to some extent, have alleviated the need for single-modal dense annotations in the 3D domain.



However, multimodal learning frameworks in this direction commonly face the challenge of assembling scalable, high-quality, and well-aligned multimodal data for 3D applications.
We identify the language modality for 3D data as the critical bottleneck in this process. Existing frameworks tend to utilize manually annotated category names and short descriptions derived from metadata as the language counterparts for the 3D data.
Those approaches~\citep{qi2023contrast,xue2022ulip}, however, lack scalability as they always rely on some degree of human annotations during the dataset collection process, which will be hard to scale up.
Furthermore, existing methods are not comprehensive enough as the derived language information might not provide sufficient details and lacks variations, or appears to be noisy. This highlights the need for an innovative paradigm to provide language counterparts for 3D data that are both scalable and comprehensive, thus truly unleashing the potential of multimodal learning.

\begin{figure*}[t]
    \centering
    \includegraphics[width=1.0\linewidth]{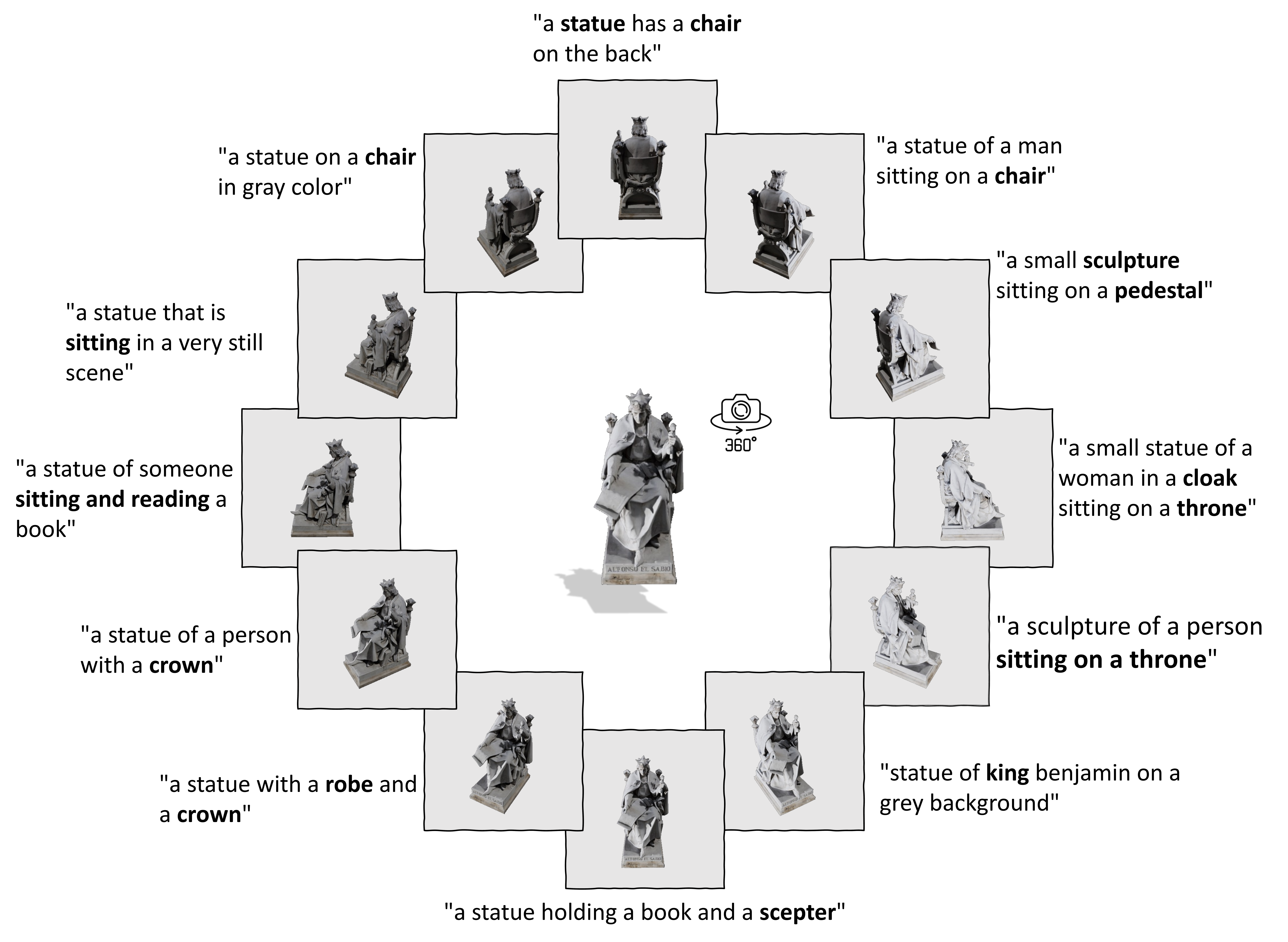}
    \caption{An illustration of language description generation from 2D images. These images are rendered from a set of holistic viewpoints of a 3D object. In some views, the chair is not visible, while in other views, the scepter/sword cannot be seen. Combining descriptions of all views is essential for the model to learn comprehensive and holistic information about the 3D object. From the metadata, the manual caption for this object is “Estatua de Alfonso X - José Alcoverro (1892)“, which doesn’t include much semantic information and could potentially harm the multimodal pre-training, unlike ULIP-2's holistic captions.}
    \label{fig:differentangle}

\end{figure*}

However, the optimal way to acquire and utilize language data for 3d modality is unclear. Although well-trained human annotators could potentially provide detailed language descriptions of 3D objects, such a method is both costly and lacks scalability. Moreover, identifying the appropriate language counterpart modality for a 3D shape is not a straightforward task.

To address these issues, we first reconsider what the 2D image counterpart modality for a 3D shape should be. Semantically, if we can render 2D images of a 3D shape from any viewpoint,  the collection of all these rendered images should approximately encapsulate all information about this 3D shape, thus forming an appropriate image counterpart modality for 3D. By analogy, if we can linguistically describe a 3D shape from any viewpoint,  the compilation of all these language descriptions from all perspectives should also approximately encompass all linguistically expressible information about this shape, thus forming an appropriate language modality for the 3D shape.
In practice, for efficiency, we may sample a finite fixed set of holistic viewpoints instead of "any viewpoint". If we apply the same set of viewpoints for creating the language modality as we render the images, this task naturally boils down to describing the rendered 2D image for a given viewpoint. Leveraging the advances in large multimodal models that are trained on extensive language and image data, we utilize their ability to generate detailed language descriptions for the rendered images. This method allows us to automate the process in a scalable way as it only needs 3D data itself, while the rich knowledge from the large multimodal models is also distilled into the language descriptions. As a result, this automated and scalable strategy enriches the language modality with detailed, holistic descriptions, further aiding multimodal 3D representation learning.


In light of the preceding reasoning, and also in response to the challenge of scalable and comprehensive multimodal 3D data acquisition, we introduce ULIP-2, a novel framework that encompasses an innovative approach to generate well-aligned, holistic multimodal data for 3D understanding, coupled with an efficient multimodal pre-training architecture capable of robustly aligning these multimodal data, thereby harnessing the full potential of multimodal learning.

Given a 3D shape, our initial step involves extracting 3D point cloud data to serve as the 3D modality input. We then render this shape into a series of images from a fixed set of holistic viewpoints, providing the 2D modality input. For each rendered image, we employ a large multimodal model to generate a list of detailed descriptions, thereby establishing the language modality (as illustrated in Figure \ref{fig:differentangle}). This approach allows us to create scalable multimodal data for 3D, as it only requires the 3D data itself. Furthermore, by generating descriptions from a comprehensive set of holistic views, we address the prior issues of detail and comprehensiveness in the language modality. By employing an efficient multimodal pre-training architecture to align this multimodal data, we facilitate the learning of a comprehensive multimodal 3D representation, as described in Figure \ref{fig:overview}. Consequently, ULIP-2 offers a promising solution for scalable and comprehensive multimodal pre-training for 3D representation learning.

ULIP-2 advances beyond its predecessor, ULIP, by (1) proposing a manual-effort-free data creation paradigm for comprehensive multimodal learning, (2) leveraging this scalable paradigm to extend multimodal 3D learning to larger datasets, while scaling up both the vision-language and 3D backbones and (3) when pre-trained on the same datasets, ULIP-2 delivers impressive improvements over ULIP on all downstream tasks.

Our paper has the following main contributions:

\begin{enumerate}
\item ULIP-2 facilitates scalable multimodal pre-training without the need for human annotations, making it applicable to any 3D dataset, even unlabeled. It relies solely on the 3D data, enabling broader applicability and ease of use.
\item ULIP-2 achieves significant advancements in multimodal representation learning. On the challenging open-world Objaverse-LVIS benchmark, ULIP-2 attains a top-1 accuracy of \textbf{50.6\%}, surpassing current SOTA (OpenShape \citep{liu2023openshape}) by significant \textbf{3.8\%}, despite ULIP-2 has a simpler and more streamlined framework; for zero-shot classification on ModelNet40, ULIP-2 reaches \textbf{84.7\%}, even outperforming some fully supervised 3D classification methods \citep{wu20153d}. Furthermore, it secures an overall accuracy of \textbf{91.5\%} on the ScanObjectNN benchmark with only 1.4 million parameters. The ULIP-2 encoder's 3D to language generation capabilities with LLMs are also demonstrated, highlighting its potential to keep pace with the growing LLM development. Moreover, ULIP-2 can effectively synergize with the ever-increasing capacity of 3D data and the development of large multimodal models.
\item We release two large-scale tri-modal datasets, \textbf{"ULIP-Objaverse"} and \textbf{"ULIP-ShapeNet"} triplets, consisting of point clouds, images, and language descriptions. The statistics of the datasets are detailed in Table \ref{table:triplet_stats}.
\end{enumerate}

\section{Related Work}
\vspace{-2mm}

\noindent\textbf{Multimodal Representation Learning}. In recent years, multimodal representation learning has emerged as a popular research topic due to its remarkable capabilities and applications. Most research works focus on learning multimodal representation for only two modalities:  language and image modalities, which have led to remarkable outcomes. One line of research in this area emphasizes the interaction between image regions and caption tokens using Transformer-based architectures \cite{huang2021seeing,kim2021vilt,li2021align,zeng2021multi, panagopoulou2023x}, which exhibit strong predictive capabilities but are computationally expensive to train. Alternatively, methods such as CLIP \cite{radford2021learning} and SLIP \cite{mu2022slip} target generating single features for image and text independently and subsequently align these two modalities. This simplified architecture promotes robust and efficient large-scale pre-training, even on noisy data.


Recent works have demonstrated promising results by extending multimodal representation learning to 3D modality. ULIP \cite{xue2022ulip} is one of the pioneering works in creating (3D point cloud - image - language) triplets, By aligning these three modalities together, ULIP enhances 3D representation learning and mitigates the need for single-modal dense 3D data annotations, thereby partially alleviating the data scarcity issue in 3D.
A recent work \cite{zhang2022learning} seeks to learn 3D representations from pre-trained 2D encoders via Image-to-Point Masked Autoencoders. However, this approach does not involve alignment with the language modality, which potentially limits its capacity for more complex multimodal tasks.
The concurrent work of ~\citep{liu2023openshape} further extends ULIP's framework to achieve stronger performance, but it still relies on manual annotation of 3D data and a complicated data engineering framework. However, ULIP-2 demonstrates that, with the proposed much simpler and streamlined framework, it can still achieve SOTA results on the challenging Objaverse-LVIS benchmark and outperform OpenShape by an impressive 3.8\% in Objaverse-LVIS top-1 accuracy.


Despite the development of methods such as ULIP to reduce the single-modal dense annotation effort, they ~\citep{xue2022ulip,qi2023contrast,liu2023openshape} still confront scalability challenges due to their dependency on dataset metadata and category names for obtaining the language counterpart modality. Additionally, the prompt-based pseudo-captions generated by these methods lack the fine-grained details, and variations that are necessary for comprehensive understanding. In contrast, ULIP-2 overcomes these limitations by leveraging the power and knowledge of state-of-the-art large multimodal models. This approach fundamentally diminishes data requirements and enriches the pre-train multimodal data, thereby enabling more efficient applications on larger datasets and yielding much stronger 3D representations.

\noindent\textbf{Generative Large Multimodal Models}. The expansion of transformer models, from GPT to GPT-4 \cite{vaswani2017attention,openai2023gpt4}, demonstrates the effectiveness of scale in multimodal tasks. This approach, originating from \cite{anderson2018bottom}, has seen considerable advancements in text generation from images \cite{lu2018neural, zhou2019grounded, zhou2020unified, cho2021unifying, wang2021simvlm, li2021align, li2020unimo, li2022blip, li2023blip}. Our study leverages BLIP-2 \cite{li2022blip} to generate diverse annotations for 3D shapes, facilitating learning richer multimodal 3D representations. We also conduct an ablation study in \ref{sec:ablate_large_multimodal_models} on the used large multimodal models which indicates that, ULIP-2 benefits from the advancements in large multimodal models, synergizing with the rapid improvements in the field.

\begin{table*}[h]
    \centering
    \small
    \begin{tabular}{lllccccc}
         \toprule
         \multirow{2}*{\textbf{Model}} & \textbf{Pre-train} & \textbf{Pre-train} & \textbf{Manual} & \multicolumn{2}{l}{\textbf{Objaverse-LVIS}} & \multicolumn{2}{l}{\textbf{ModelNet40}}\\
         & \textbf{dataset} & \textbf{method} & \textbf{captions?} & \textbf{top-1} & \textbf{top-5} & \textbf{top-1} & \textbf{top-5}  \\
         \midrule
         
         PointCLIP ~\citep{zhang2022pointclip} & -- & -- & -- & 1.9 & 5.8 & 19.3 & 34.8 \\
         PointCLIPv2 ~\citep{Zhu2022PointCLIPV2} & -- & -- & -- & 4.7 & 12.9 & 63.6 & 85.0 \\
         ReCon ~\citep{qi2023contrast} & ShapeNet & ReCon ~\citep{qi2023contrast} & \cmark  & 1.1 & 3.7 & 61.2 & 78.1 \\
         CLIP2Point ~\citep{huang2022clip2point} & ShapeNet & CLIP2Point ~\citep{huang2022clip2point} & \xmark & 2.7 & 7.9 & 49.5 & 81.2 \\
         
         \rowcolor{gray!10} Point-BERT ~\citep{yu2022point} & ShapeNet & OpenShape ~\citep{liu2023openshape} & \cmark & 10.8 & 25.0 & 70.3 & 91.3 \\
         \addlinespace
         \rowcolor{gray!10} Point-BERT ~\citep{yu2022point} & Objaverse(no LVIS) + ShapeNet & OpenShape ~\citep{liu2023openshape} & \cmark & 38.8 & 68.8 & 83.9 & 97.6 \\
         \addlinespace
         \rowcolor{gray!10} Point-BERT ~\citep{yu2022point} & Objaverse + ShapeNet & OpenShape ~\citep{liu2023openshape} & \cmark & 46.5 & 76.3 & 82.6 & 96.9 \\
         \addlinespace
         \rowcolor{gray!10} Point-BERT ~\citep{yu2022point} & Objaverse + ShapeNet + (2 extra) & OpenShape ~\citep{liu2023openshape} & \cmark & 46.8 & 77.0 & 84.4 & \textbf{98.0} \\

         \midrule
         
         \multirow{6}*{Point-BERT ~\citep{yu2022point}} & \multirow{2}*{ShapeNet} & ULIP \cite{xue2022ulip} & \cmark & 2.6 & 8.1 & 60.4 & 84.0  \\

         & & ULIP-2 & \xmark & 16.4 & 34.3 & 75.2 & 95.0  \\
         
         \cmidrule(lr){2-8}

         & \multirow{2}*{Objaverse(no LVIS) + ShapeNet}
         & ULIP \cite{xue2022ulip} & \cmark & 21.4 & 41.9 & 68.6 & 86.4 \\
         
         & & ULIP-2 & \xmark & 46.3 & 75.0 & 84.0  & 97.2  \\

         \cmidrule(lr){2-8}

         & \multirow{2}*{Objaverse + ShapeNet}
         & ULIP \cite{xue2022ulip} & \cmark & 34.9 & 61.0 & 69.6 & 85.9 \\
         & & ULIP-2 & \xmark & \textbf{50.6} & \textbf{79.1}  & \textbf{84.7}  & 97.1  \\
         
         \bottomrule
    \end{tabular}
    \caption{Zero-shot 3D classification on Objaverse-LVIS and ModelNet40. The highlighted lines of OpenShape are from the current SOTA approach. Our method surpasses the current state-of-the-art (SOTA) OpenShape in zero-shot 3D classification, achieving a 3.8\% higher top-1 accuracy on Objaverse-LVIS, and demonstrating comparable performance on ModelNet40, despite using fewer pre-training datasets. A tick in the “Manual captions?” column means the pre-trained model leverages 3D captions that, to some degree, rely on manual efforts, while a cross means the opposite.}
    \label{tab:zero-shot-lvis-modelnet}
\end{table*}

\vspace{-2mm}
\noindent\textbf{3D Point Cloud Understanding}.
PointNet~\citep{qi2017pointnet} is a pioneering work that processes 3D point clouds directly~\citep{qi2017pointnet++}. Building on this, PointNeXt~\citep{qian2022pointnext} emerges as a light-weight, high-performance variant. In the realm of self-supervised pre-training for point clouds, Point-BERT~\citep{yu2022point} moves a significant step forward with its transformer-based architecture, showcasing notable performance in zero-shot classification tasks. In ULIP-2, we leverage both Point-BERT and PointNeXt as our 3D encoders to harness their strong capabilities.

\section{Method}


ULIP-2 assimilates the pre-training framework of ULIP and introduces a scalable and comprehensive multimodal triplet creation paradigm that not only eliminates the need for human annotations but also significantly improves the learned multimodal 3D representations. By merging ULIP’s efficient multimodal pre-training with this scalable triplet creation method, ULIP-2 paves the way for large-scale pre-training that essentially operates in a pseudo-self-supervised manner. We demonstrate that this method effectively mitigates the data scalability issue, and simultaneously advances the field of 3D representation learning to a new level of performance.


\subsection{Preliminary: ULIP}

ULIP \cite{xue2022ulip} presents an efficient multimodal pre-training framework that constructs triplets encompassing three modalities: (1) the 3D modality, obtained by extracting 3D point cloud data; (2) the image modality, generated by rendering images from 3D shapes across multiple viewpoints; and (3) the language modality, derived by prompting dataset metadata such as descriptive terms and category names into cohesive sentences.

ULIP utilizes the ViT-B encoders from SLIP \cite{mu2022slip}, a pre-trained vision-language model and a variant of the CLIP model, to learn 3D representations. It accomplishes this by aligning 3D modality features to the feature space shared by language and image modalities. ULIP-2 shares a similar objective with ULIP in aligning the (image, text, 3D) modalities, which prompts us to adopt its pre-training framework. Given the close resemblance in setup between ULIP and ULIP-2, we choose ULIP as our experimental baseline.


\subsection{Scalable Triplet Creation}

In ULIP-2, the model similarly utilizes three input modalities, though it only requires the 3D object data itself. As depicted in Fig. \ref{fig:overview}, given a 3D object, we extract 3D point clouds from the surface as the input to the 3D encoder and generate images from various viewing angles. We then leverage BLIP-2 \cite{li2023blip}, a cutting-edge large multimodal model, to generate descriptive texts for each rendered 2D image. For each image, we generate a set of sentences, rank them using CLIP similarities, and aggregate the top-1 description to form the language modality in the triplet.

This scalable triplet creation approach facilitates dataset scaling, eliminating the need for dataset metadata collection and necessitating only the 3D data itself. Our method is capable of aligning 3D representations with holistic image-text pairs in any unannotated 3D dataset, thereby providing a more scalable and comprehensive solution.


\subsection{Tri-modal Pre-training}
ULIP-2 aligns the triplet of 3D point clouds, 2D rendered images, and comprehensive descriptions to a unified feature space. We adopt the largest version of encoders from OpenCLIP (ViT-G/14) \cite{ilharco_gabriel_2021_5143773} for most of our experiments and freeze it during the pre-training. The feature space, already pre-aligned by OpenCLIP, serves as the target space where we aim to integrate the 3D modality.

During tri-modal pre-training, given a 3D shape $\mathbf{O}$, we extract its 3D point cloud $\mathbf{P}$, randomly sample its 2D rendered image $\mathbf{I}\sim\mathsf{render}(\mathbf{O})$, with its BLIP-2 generated language description $\mathbf{T}\sim\mathsf{blip2}(\mathbf{I})$, where $\mathsf{render}$ is the 3D-to-2D rendering operation and $\mathsf{blip2}$ is to query BLIP-2 \cite{li2023blip} for image description. We then extract the image feature $\mathbf{f}^\text{I}=E_\text{I}(\mathbf{I})$ and text feature $\mathbf{f}^\text{T}=E_\text{T}(\mathbf{T})$ based on the pre-aligned and frozen image encoder $E_\text{I}$ and text encoder $E_\text{T}$ in OpenCLIP \cite{ilharco_gabriel_2021_5143773}. We target to train a 3D point cloud encoder $E_\text{P}$ such that its 3D feature $\mathbf{f}^\text{P}=E_\text{P}(\mathbf{P})$ is aligned with its image and text features. We formulate the 3D-to-image alignment using the contrastive loss similar in spirit to CLIP \cite{radford2021learning}:
\begin{equation}
\mathcal{L}_\text{P2I}=-\frac{1}{2}\sum_i\log\frac{\exp(\mathbf{f}_i^\text{P} \mathbf{f}_i^\text{I}/\tau)}{\sum_j\exp(\mathbf{f}_i^\text{P} \mathbf{f}_j^\text{I}/\tau)}+\log\frac{\exp(\mathbf{f}_i^\text{P} \mathbf{f}_i^\text{I}/\tau)}{\sum_j\exp(\mathbf{f}_j^\text{P} \mathbf{f}_i^\text{I}/\tau)},
\end{equation}
where $i$, $j$ are the sampling indices, and $\tau$ is a learnable temperature parameter. The first term indicates that the dot product of the 3D feature and the image feature of the same sample should stand out among other products where the \textit{image features} are from different samples. Likewise, the second term indicates that the dot product of the 3D feature and the image feature of the same sample should stand out among other products where the \textit{3D features} are from different samples.

Similarly, we formulate the 3D-to-text alignment loss as:
\begin{equation}
\mathcal{L}_\text{P2T}=-\frac{1}{2}\sum_i\log\frac{\exp(\mathbf{f}_i^\text{P} \mathbf{f}_i^\text{T}/\tau)}{\sum_j\exp(\mathbf{f}_i^\text{P} \mathbf{f}_j^\text{T}/\tau)}+\log\frac{\exp(\mathbf{f}_i^\text{P} \mathbf{f}_i^\text{T}/\tau)}{\sum_j\exp(\mathbf{f}_j^\text{P} \mathbf{f}_i^\text{T}/\tau)}.
\end{equation}

Our final training objective is to train the 3D encoder $E_\text{P}$ that minimizes the sum of the two contrastive alignment losses above:
\vspace{-2mm}
\begin{equation}
\min_{E_\text{P}}\mathcal{L}_\text{P2I}+\mathcal{L}_\text{P2T}.
\end{equation}
\vspace{-6mm}

\subsection{Scaling Up the 3D Multimodal Learning}

Recognizing the benefits of more powerful image and text encoders for learning more generalized multimodal 3D representations, we extend our exploration beyond the smaller ViT-B model, previously utilized in ULIP. Our experiments focus on upgrading this vision-language backbone in the tri-modal alignment framework. Additionally, we investigate scaling up the model size, while keeping the other settings unchanged. The effectiveness of these modifications is evaluated through zero-shot classification tasks on both ModelNet40 and Objaverse-LVIS datasets. See Table \ref{tab:ablate_scale_up}.

\section{Experiments}

\subsection{ULIP-Objaverse Triplets and ULIP-ShapeNet Triplets Creation}
We extract triplets of 3D point clouds, images, and language descriptions based on two large-scale datasets of 3D shapes. The first dataset is Objaverse \cite{Deitke2022objaverse}, the recently released and largest-scale realistic 3D dataset. It has $\sim$ 800K real-world 3D shapes, each of which is associated with metadata containing a "name" field.  For each 3D shape, we use Blender \cite{kent20153d} to render 12 images, spaced equally by 360/12 degrees. For each rendered image, we employ BLIP-2-opt6.7B in BLIP-2~\citep{li2023blip} to generate 10 detailed descriptions independently, which are then ranked using CLIP-VIT-Large~\citep{radford2021learning} image-text similarity score. Based on an ablation study in Sec. \ref{sec:top_k_captions}, we use the top 1 description as the language modality input. Following ULIP and OpenShape, we use 10k, 8k, and 2k points from each 3D shape to accommodate different downstream tasks. Our generated well-paired triplets of comprehensive descriptions, 2D rendering images, and 3D point clouds are released as \textbf{ULIP-Objaverse} triplets.

The second dataset is ShapeNet~\citep{chang2015shapenet}, a renowned synthetic dataset. We employ its publicly available subset which has $\sim$ 52.5K 3D shapes with 55 annotated categories. For each shape, we follow ULIP to sample 30 equally spaced view angles, for each view angle, we render an RGB image and a depth map. The image description generation method is the same as that in Objaverse. We release these triplets as \textbf{ULIP-ShapeNet} triplets. More implementation details and ablation studies are included in the Appendix.

\begin{table}[t]
  \small
  \centering
  \begin{tabular}{lcc}
    \toprule
    \textbf{Modality} & \textbf{ULIP-Objaverse} & \textbf{ULIP-ShapeNet} \\
    \midrule
    Point Clouds & $\sim$ 800k & $\sim$ 52.5k \\
    Images & $\sim$ 10 million & $\sim$ 3 million \\
    Language & $\sim$ 100 million & $\sim$ 30 million \\
    \bottomrule
  \end{tabular}
  \caption{Statistics of ULIP-Objaverse and ULIP-ShapeNet triplets.}
    \label{table:triplet_stats}
    \vspace{-4mm}
\end{table}

\subsection{Downstream Tasks}

We use the ModelNet40~\citep{chang2015shapenet}, Objaverse-LVIS~\citep{Deitke2022objaverse}, and ScanObjectNN~\citep{Uy_2019_ICCV} datasets to benchmark ULIP-2. ModelNet40 is a synthetic CAD model dataset. It contains $\sim$ 9.8k training samples and $\sim$ 2.5k testing samples. Objaverse-LVIS is a subset of the Objaverse dataset with human-verified category labels. It has $\sim$ 46k samples spanning $\sim$ 1.2k categories, which is suitable for more challenging open-world zero-shot 3d shape classification. ScanObjectNN is a real-world 3D dataset with $\sim$ 2.9k samples under 15 categories. We follow the same dataset setup and preparation protocols used in ULIP and OpenShape, ensuring consistency in our comparisons.

We conduct experiments on three downstream tasks: (1) the zero-shot 3D classification task involving multimodal inputs and (2) the standard 3D classification task involving a single modality and (3) the 3D captioning task involving 3D-to-language generation with LLMs.

\noindent\textbf{Evaluation Metrics}. We adopt the same evaluation metrics used in ULIP: top-1 and top-5 accuracy for the zero-shot 3D classification task; overall accuracy and class average accuracy for the standard 3D classification task. For the new downstream task, 3D-to-language generation, we follow X-InstructBLIP~\citep{xinstructblip} and employ CIDEr ~\citep{vedantam2015cider} score to quantitatively evaluate the quality of generated captions.

\noindent\textbf{Backbones}. We pre-train ULIP-2 on two representative backbones: Point-BERT~\citep{yu2022point} is a transformer-based backbone that performs strongly in ULIP's zero-shot classification experiments. PointNeXt \cite{qian2022pointnext} is a work that proposes a lightweight backbone based on PointNet++ \cite{qi2017pointnet++} and delivers promising results on the ScanObjectNN benchmark.

\subsection{Comparisons to Baselines}

\noindent\textbf{Zero-Shot 3D Classification}. We follow the same procedure as in ULIP and OpenShape for zero-shot 3D classification, and compare with existing zero-shot approaches, including ~\citep{xue2022ulip, liu2023openshape, zhang2022pointclip, Zhu2022PointCLIPV2, qi2023contrast, huang2022clip2point}. We present the zero-shot 3D classification results on both Objaverse-LVIS and ModelNet40 in Table \ref{tab:zero-shot-lvis-modelnet}. First, we observe that, when \textbf{pre-trained on the same datasets}, benefit from ULIP-2 pre-training, Point-BERT obtains significantly better results than those pre-trained with ULIP. Specifically, when pre-train both ULIP and ULIP-2 on ShapeNet, ULIP-2 outperforms ModelNet40 top-1 accuracy over ULIP by \textbf{14.8\%} and outperforms Objaverse-LVIS top-1 accuracy by \textbf{13.8\%}. If pre-trained on Objaverse (excluding LVIS samples) and ShapeNet jointly, ULIP-2 outperforms ULIP by \textbf{15.4\%} on ModelNet40 top-1 accuracy and outperforms ULIP by \textbf{24.9\%} on Objaverse-LVIS top-1 accuracy. These gains underscore the efficacy of our approach, particularly the ranked holistic-view language descriptions and scaling strategies that amplify pre-training representation capabilities. The comprehensive language descriptions generated by the large multimodal model encapsulate its knowledge from a vast amount of language and image data, thus enriching the semantic richness of 3D shape descriptions and bolstering the alignment between language and 3D modalities.

Moreover, with ULIP-2's simple and streamlined framework, it achieves performance outperforming pre-existing baselines including concurrent OpenShape~\citep{liu2023openshape}, which is also the current SOTA (46.8\% in Objaverse-LVIS top-1).

\begin{figure}[t]
    \centering
    \includegraphics[width=0.8\linewidth]{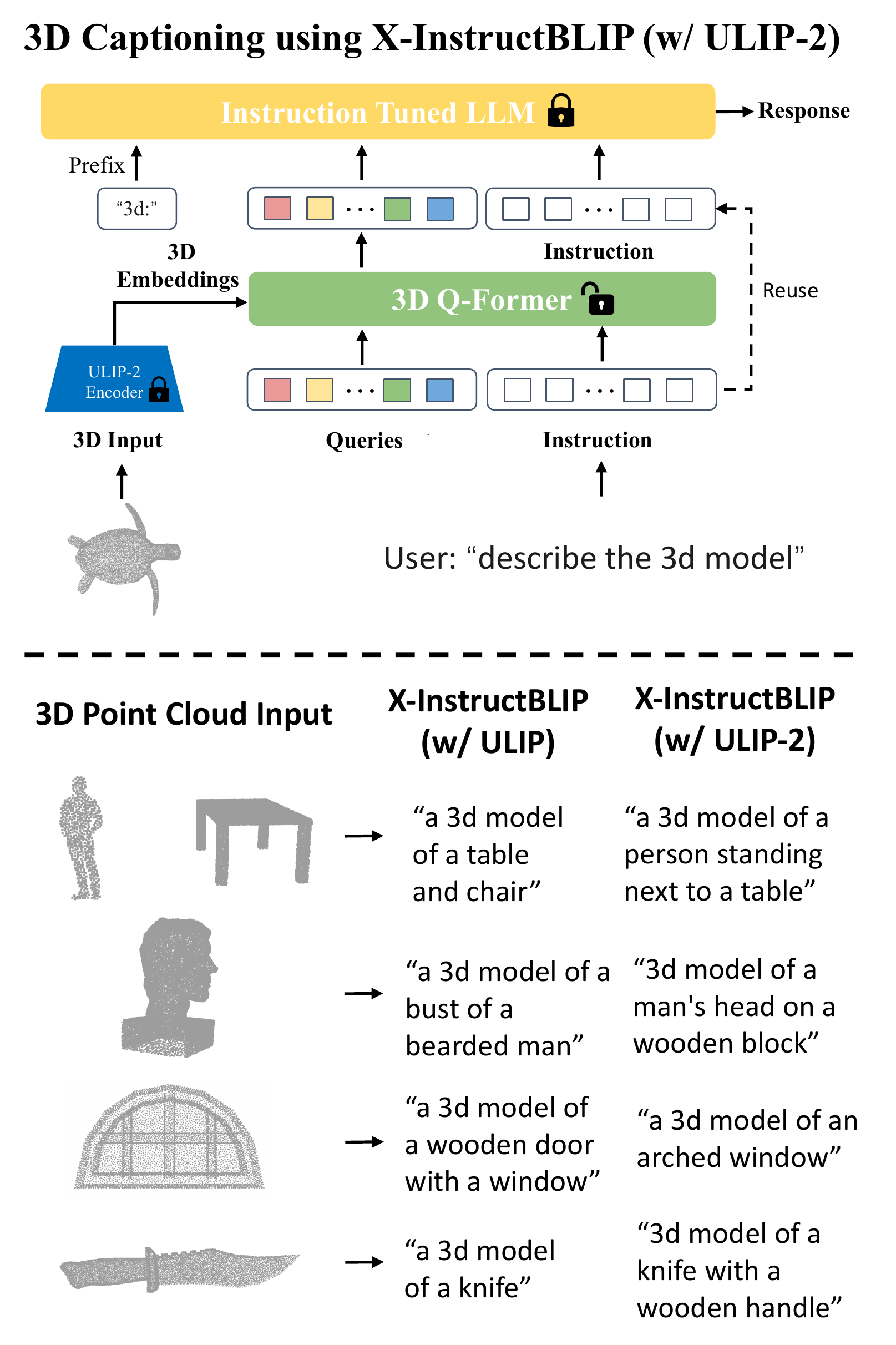}
    \caption{3D-to-language multimodal generation using X-InstructBLIP framework~\citep{xinstructblip}.}
    
    \label{fig:xinstructblip}
\end{figure}

\begin{table}[htb]
    \centering
    \resizebox{\linewidth}{!}{
    \begin{tabular}{lccc}
    \toprule
          \multirow{2}*{\textbf{Model}} & \textbf{\#Params} & \textbf{Overall} & \textbf{Class-average} \\
          &  \textbf{(M)} & \textbf{accuracy} & \textbf{accuracy} \\
         \midrule
         PointNet \cite{qi2017pointnet} &  3.5 & 68.2 & 63.4 \\
         PointNet++ \cite{qi2017pointnet++} &  1.5 & 77.9 & 75.4 \\
         DGCNN \cite{wu2018dgcnn} & 1.8 & 78.1 & 73.6 \\
         MVTN \cite{hamdi2021mvtn} &  11.2 & 82.8 &  --\\
         RepSurf-U \cite{ran2022surface} &  1.5 & 84.6 &  --\\
         Point-MAE \cite{pang2022masked} &  22.1 & 85.2 & -- \\
         PointMLP \cite{ma2022rethinking} & 12.6 &  85.7 & 84.4 \\
         Point-M2AE \cite{zhang2022point} &  15.3 & 86.4 & -- \\
         PointCMT \cite{yan2022let} & 12.6 & 86.7 & 84.8 \\
         ACT \cite{dong2022autoencoders} & 22.1 & 88.2 & --\\
         P2P \cite{wang2022p2p} &  - & 89.3 & -- \\
         Recon-s \cite{qi2023contrast} &  19.0 & 89.5 & -- \\
         I2P-MAE \cite{zhang2022learning} &  12.9 & 90.1 & -- \\
         \midrule
         Point-BERT (official) & 22.1 & 83.1 &  --\\
         Point-BERT (w/ ULIP) & 22.1 & 88.7 & --\\
         Point-BERT (w/ ULIP-2)  & 22.1 & 89.7 
         & --\\
         \midrule
         PointNeXt (from scratch)  & 1.4 & 87.5 & 85.9 \\
         PointNeXt (w/ ULIP)  & 1.4 & 90.1 & 89.2 \\
         PointNeXt (w/ ULIP-2) & 1.4 & 91.1
         & 90.3
         \\
         PointNeXt (w/ ULIP-2)*  & 1.4 & \textbf{91.5} 
         & \textbf{90.9}
         \\
         \bottomrule
    \end{tabular}}
    \caption{3D classification results on ScanObjectNN. ULIP-2 significantly outperforms the baselines. * means the voting \cite{ma2022rethinking} is used.} 
    \label{tab:fintune-scan}
    \vspace{-4mm}
\end{table}

\noindent\textbf{Standard 3D Classification}. We adhere to ULIP and community protocols for standard 3D classification. We present 3D classification results on ScanObjectNN hardest set in Table \ref{tab:fintune-scan}. When pre-trained on Objaverse and ShapeNet jointly for the same 3D encoder architecture with both ULIP and ULIP-2 frameworks, we observe that ULIP-2 (using the Point-BERT backbone) improves the baseline method (without multimodal pre-training) by 6.6\%. Using the PointNeXt backbone, ULIP-2 achieves a significant 4.0\% performance gain over training from scratch, achieving an overall accuracy of 91.5\% and establishing a new record on the ScanObjectNN benchmark with just 1.4 million parameters.

\noindent\textbf{3D-to-Language Generation}. As depicted in Figure \ref{fig:xinstructblip}, we adopt the X-InstructBLIP methodology~\citep{xinstructblip} to integrate the ULIP-2 pre-trained encoder with a frozen large language model (LLM), endowing it with the capability to generate language from 3D data. For a fair comparison of 3D-to-language generation abilities, we plug frozen Point-BERT models into X-InstructBLIP's ~\citep{xinstructblip} framework, which are pre-trained under both ULIP and ULIP-2 frameworks with the same pre-training datasets (Objaverse + ShapeNet). Then we benchmark the 3D captioning abilities following ~\citep{xinstructblip}. All other variables are held constant during this evaluation. Captioning performance is measured using the PyCOCOTools Cider Score~\citep{vedantam2015cider}, offering a quantitative analysis of the models' captioning performance. Table \ref{tab:3d_captioning} shows that ULIP-2 pre-trained encoder can significantly improve the captioning score by 28.3\%, and Figure \ref{fig:xinstructblip} shows qualitatively the generated captions using ULIP-2 pre-trained model is more accurate and descriptive.

\begin{table}[h]
    \centering
    \vspace{-3mm}
    \small
    \begin{tabular}{ccc}
        \toprule
        \textbf{Multimodal} & \textbf{Frozen 3D} & \textbf{CIDEr} \\
        \textbf{generation framework} & \textbf{encoder} & \textbf{score} \\
        \midrule
        X-InstructBLIP & PB w/ ULIP & 132.2 \\
        X-InstructBLIP & PB w/ ULIP-2 & \textbf{160.5}
        \\
        \bottomrule
    \end{tabular}
    \caption{3D-to-language generation using X-InstructBLIP ~\citep{xinstructblip}, pre-trained on the same Objaverse + ShapeNet datasets setting. PB w/ ULIP-2 means Point-BERT pre-trained with ULIP-2 framework.}
    \label{tab:3d_captioning}
    \vspace{-2mm}
\end{table}

\vspace{-6mm}
\section{Ablation Study}

\subsection{Ablation on the effect of the generated captions}
\label{sec:ablate_captions}
To ablate how the generated captions contribute to the performance, we conducted experiments aligned with the ULIP's settings, but with only one key modification: the language modality. Instead of using ULIP's manual descriptions, we utilized the top-1 ranked holistic-view captions generated by BLIP-2. Results in Table \ref{tab:ablate-captions} show significant improvements in zero-shot classification on ModelNet40 when using these generated captions, demonstrating their crucial impact compared to the manual captions used in ULIP.

\begin{table}[h!]
    \centering
    \small
    \begin{tabular}{ccc}
         \toprule
         \textbf{Pre-train} & \multicolumn{2}{l}{
         \textbf{ModelNet40}}\\
         \textbf{language modality} & \textbf{top-1} & \textbf{top-5}\\
         \midrule
         
         Manual captions & 60.4 &  84.0\\
         Top-1 holistic BLIP-2 captions  & \textbf{69.7}
         & \textbf{88.1}
         \\

         \bottomrule
    \end{tabular}
    \caption{Point-BERT zero-shot 3D classification on ModelNet40, pre-trained on ShapeNet with SLIP ViT-B encoders (used in ULIP).}
    \label{tab:ablate-captions}
\end{table}

\subsection{Different Large Multimodal Models}
\label{sec:ablate_large_multimodal_models}
Considering that the language description quality from large multimodal models plays an important role in 3D representation pre-training, we conduct an ablation study over two such models. We use BLIP-2 ~\citep{li2023blip} throughout the benchmarking experiments above. We hereby compare it to its earlier version BLIP~\cite{li2022blip} for the zero-shot 3D classification task using Point-BERT backbone pre-trained on ShapeNet. Results in Table \ref{tab:ablate_large_multimodal_model} show that using BLIP-2 generated descriptions achieves better results than BLIP, thanks to its evolved image understanding capability, suggesting that as the large multimodal models advance, the performance of ULIP-2 can be expected to improve correspondingly.

\begin{table}[h!]
    \centering
    \vspace{-2mm}
    \begin{tabular}{lcc}
        \toprule
        \textbf{Large multimodal models}  & \textbf{top-1} & \textbf{top-5}
        \\
        \midrule
        BLIP \cite{li2022blip} & 67.7 & 88.6 \\
        BLIP-2 \cite{li2023blip} & \textbf{69.7} & \textbf{88.8}\\
        \bottomrule
    \end{tabular}
    \caption{Zero-shot 3D classification on ModelNet40. Pre-trained on ShapeNet with SLIP ViT-B encoders.}
    \label{tab:ablate_large_multimodal_model}
    \vspace{-4mm}
\end{table}



\subsection{Number of 2D Views Per 3D Object}
We further perform an ablation study for the zero-shot 3D classification performance w.r.t. the number of holistic views w/ its top-1 BLIP-2 caption in pre-training. Results in Table \ref{tab:ablate_num_views} demonstrate that, with the increase in the number of views, zero-shot classification accuracy increases accordingly. This validates our statement that diverse language descriptions of holistic views benefit multimodal 3D representation learning.


\begin{table}[htbp]
    \centering
    \vspace{-2mm}
    \begin{tabular}{ccc}
        \toprule
                     \multirow{2}*{\textbf{\# Holistic views}} & \multicolumn{2}{c}{\textbf{Accuracy}}
        \\
         & \textbf{top-1} & \textbf{top-5}
        \\
        \midrule
         1 & 54.8 & 77.9 \\
         2 & 58.1 & 80.5 \\
         15 & 69.3 & 88.6 \\
         30 & \textbf{69.7} & \textbf{88.8} \\
        \bottomrule
    \end{tabular}
    \caption{Zero-shot 3D classification on ModelNet40, pre-trained on ShapeNet with SLIP ViT-B encoders.}
    \label{tab:ablate_num_views}
    \vspace{-4mm}
\end{table}

\vspace{-2mm}
\subsection{Top-\texorpdfstring{$k$}{k} CLIP Ranked Captions Per 2D View}
\label{sec:top_k_captions}
To assess the effectiveness of our top-1 BLIP-2 caption strategy, we conducted an ablation study on selecting different top-k of the 10 independently-generated BLIP-2 captions for the multimodal pre-training. The results in Table \ref{tab:ablate_top_k} indicate that, using the top-1 CLIP score ranked caption yields the best performance. This makes intuitive sense: the top-1 CLIP-scored caption tends to be more noise-proof, which is advantageous for multimodal learning in our context.

\begin{table}[h]
    \centering
    \begin{tabular}{ccc}
        \toprule
        \multirow{2}*{\textbf{Top-\(k\) BLIP-2 captions selected}} & \multicolumn{2}{c}{\textbf{Accuracy}}
        \\
        & \textbf{top-1} & \textbf{top-5}
        \\
        \midrule
        1 & \textbf{69.7} & \textbf{88.8} \\
        3 & 66.7 & 87.2 \\
        5 & 66.4 & 87.7 \\
        10 & 66.3 & 85.1 \\
        \bottomrule
    \end{tabular}
    \caption{ULIP-2 zero-shot 3D classification on ModelNet40, pre-trained on ShapeNet with SLIP ViT-B. For example, top-5 BLIP-2 captions selected means that in the pre-training, we will ensemble the top-5 CLIP CLIP ranked captions as the language modality.}
    \label{tab:ablate_top_k}
    \vspace{-2mm}
\end{table}


\subsection{Scaling Up the Backbone Models}
\label{sec:ablate_scaling}

We examined the effectiveness of increasing the size of both the CLIP model and the 3D backbone model on performance. As Table \ref{tab:ablate_scale_up} illustrates, a larger CLIP model improves results. For the 3D backbone, performance peaks at around 32.5M parameters, beyond which gains diminish. Therefore, this configuration is our chosen setup, balancing between performance and model size.

\begin{table}[h]
    \small
    \centering
    \vspace{-2mm}
    \begin{tabular}{cccccc}
         \toprule
        \textbf{CLIP} &\textbf{3D encoder }  & \multicolumn{2}{c}{\textbf{ModelNet40}} & \multicolumn{2}{c}{\textbf{Objaverse-LVIS}}\\
        \textbf{size} & \textbf{\#Params(M)} & \textbf{top-1} & \textbf{top-5}  & \textbf{top-1} &\textbf{top-5}  \\
         \midrule
         ViT-B  & 21.9 & 71.4  & 89.7 & 28.3 & 52.6 \\
         ViT-G & 21.9 & 76.3 & 94.1 & 35.0 & 62.5 \\
         \cmidrule{1-6}
          ViT-G & 5.3 & 75.0 & 94.7 & 34.1 & 61.1  \\
          ViT-G & 21.9 & 76.3 & 94.1 & 35.0 & 62.5  \\
          \rowcolor{gray!25} ViT-G & 32.5  & 77.0 & 94.0 & 35.7 & 62.9  \\
          ViT-G & 43.1  & 76.8 & 94.8  & 35.9 & 62.6  \\
          ViT-G & 85.7  & 76.5 & 94.7  & 35.9 & 62.7  \\

         \bottomrule
    \end{tabular}
    \caption{Zero-shot 3D classification on ModelNet40 and Objaverse-LVIS, all models are Point-BERT models which are pre-trained on Objaverse(no-LVIS). The highlighted gray line is the model setting we use to scale ULIP-2 to the larger Objaverse dataset. The smallest 5.3M model is used when only pre-trained on ShapeNet.}
    \label{tab:ablate_scale_up}
    \vspace{-4mm}
\end{table}

         

\vspace{-2mm}
\section{Conclusion and Discussion}
\vspace{-2mm}
We present ULIP-2, a novel framework for multimodal 3D representation learning. By leveraging large multimodal models for language description generation and scaling up the multimodal 3D pre-training, ULIP-2 not only addresses the quality and scalability challenges in existing multimodal 3D datasets but also demonstrates significant improvements in all downstream tasks. We also release "ULIP-Objaverse" triplets and "ULIP-ShapeNet" triplets, two large-scale tri-modal datasets to foster further research.

\vspace{-4mm}
\paragraph{Limitations.}
ULIP-2's pre-training primarily utilizes object-level 3D shape datasets, which inherently differ from scene-level 3D data in their distribution and complexity. Exploring the application of the ULIP-2 framework to scene-level 3D data understanding, and leveraging the knowledge learned from object-level 3D data for this purpose, represents a compelling avenue for future research.

\vspace{-4mm}
\paragraph{Broader Impact.}
ULIP-2 aims to minimize human annotation in 3D multimodal pre-training, reducing labor but potentially impacting low-skilled job markets. This dual impact, a common concern in AI advancements, underscores the need for broader considerations in AI research.

{
    \small
    \bibliographystyle{ieeenat_fullname}
    \bibliography{main}
}

\clearpage
\appendix

\maketitlesupplementary
\setcounter{page}{1}

\section{Appendix}

\label{sec:appendix}

\subsection{Ablation on 3D Input}
\label{sec:ablate_color}
In order to fairly compare to OpenShape on Objaverse-LVIS benchmark, which utilizes 10k colored point clouds as the 3D input, we adopt the same 3D input preprocessing as in OpenShape. We conduct this ablation study to assess how color information influences zero-shot classification results on Objaverse-LVIS. To do this, we evaluated the ULIP-2 pre-trained Point-BERT model, pre-trained on Objaverse + ShapeNet. Our findings from Table \ref{tab:ablate_3d_input} show that ULIP-2 maintains strong performance on Objaverse-LVIS zero-shot classification tasks, even without using color information.
\begin{table}[h!]
    \centering
    \begin{tabular}{ccc}
         \toprule
         \multirow{2}*{\textbf{3D Encoder Input}} & \multicolumn{2}{l}{\hspace{0pt}\textbf{Objaverse-LVIS}}\\
          & \textbf{top-1} & \textbf{top-5}\\
         \midrule
         
         8k xyz & 48.9 &  77.1\\
         10k xyzrgb & 50.6 & 79.1 \\

         \bottomrule
    \end{tabular}
    \caption{Point-BERT w/ ULIP-2 zero-shot 3D classification on Objaverse-LVIS, pre-trained on Obajverse and ShapeNet jointly with OpenCLIP ViT-G encoders.}
    \label{tab:ablate_3d_input}
\end{table}

\subsection{Different Kinds of 3D Backbones}
\label{sec:ablate_3d_backbone}
To verify ULIP-2's improvement is agnostic to 3D backbones, we conduct experiments on the PointNeXt backbone. Results in Table \ref{tab:ablate_different_3d_backbone} show that, with another kind of intrinsically different 3D backbone, ULIP-2 can still improve the performance significantly. Given that Point-BERT is a scale-up-friendly transformer-based architecture and has better zero-shot classification results, we mainly conduct experiments on Point-BERT for all our experiments.

\begin{table}[h!]
    \centering
    \begin{tabular}{llcc}
         \toprule
         \multirow{2}*{\textbf{Model}}  & \textbf{Pre-train} & \multicolumn{2}{l}{\textbf{ModelNet40}}\\
         & \textbf{method} & \textbf{top-1} & \textbf{top-5}\\
         \midrule
         
         \multirow{2}*{PointNeXt ~\citep{qian2022pointnext}} & ULIP \cite{xue2022ulip} & 56.2 &  77.0\\
         & ULIP-2 & 72.8
         & 95.7 
         \\
         \midrule

         \multirow{2}*{Point-BERT ~\citep{yu2022point}} & ULIP ~\citep{xue2022ulip} & 60.4 & 84.0 \\
         & ULIP-2 & \textbf{75.2} & \textbf{95.0}\\

         \bottomrule
    \end{tabular}
    \caption{Zero-shot 3D classification on ModelNet40 with different 3D backbones, pre-trained on ShapeNet with SLIP ViT-B encoders.}
    \label{tab:ablate_different_3d_backbone}
\end{table}

\end{document}